# Initialization of Self-Organizing Maps: Principal Components Versus Random Initialization. A Case Study

Ayodeji A. Akinduko[1] and Evgeny M. Mirkes[2]

[1]University of Leicester, UK, aaa78@le.ac.uk
[2]Siberian Federal University, Krasnoyarsk, Russia, mirkes@bk.ru

**Abstract**
The performance of the Self-Organizing Map (SOM) algorithm is dependent on the initial weights of the map. The different initialization methods can broadly be classified into random and data analysis based initialization approach. In this paper, the performance of *random initialization* (RI) approach is compared to that of *principal component initialization* (PCI) in which the initial map weights are chosen from the space of the principal component. Performance is evaluated by the fraction of variance unexplained (FVU). Datasets were classified into quasi-linear and non-linear and it was observed that RI performed better for non-linear datasets; however the performance of PCI approach remains inconclusive for quasi-linear datasets.

**Introduction**
Self–Organizing Map (SOM) can be considered as a non-linear generalization of the principal component analysis [14] and has found much application in data exploration especially in data visualization, vector quantization and dimension reduction. Inspired by biological neural networks, it is a type of artificial neural network which uses an unsupervised learning algorithm with the additional property that it preserves the topological mapping from input space to output space making it a great tool for visualization of high dimensional data in a lower dimension. Originally developed for visualization of distribution of metric vectors [12], SOM found early application in speech recognition.

However, like clustering algorithms, the quality of learning of SOM is greatly influenced by the initial conditions: initial weight of the map, the neighbourhood function, the learning rate, sequence of training vector and number of iterations. [1][12][11]. Several initialization approaches have been developed and can be broadly grouped into two classes: random initialization and data analysis based initialization [1]. Due to many possible initial configurations when using random approach, several attempts are usually made and the best initial configuration is adopted. However, for the data analysis based approach, certain statistical data analysis and data classification methods are used to determine the initial configuration; a popular method is selecting the initial weights from the same space spanned by the linear principal component (first eigenvectors corresponding to the largest eigenvalues of the empirical covariance matrix). Modification to the PCA approach was done by [1] and over the years other initialization methods have been proposed. An example is given by [4].

In this paper we consider the performance in terms of the quality of learning of the SOM using the *random initialization* (RI) method (in which the initial weight is taking from the sample data) and the *principal component initialization* (PCI) method. The quality of learning is determined by the fraction of variance unexplained [8]. To ensure an exhaustive study, synthetic data sets distributed along various shapes of only 2-dimensions are considered in this study and the map is 1-dimensional. 1 Dimension SOM is very important, for example, for approximation of principal curves. The experiment was performed using the PCA, SOM and GSOM applet available online [8].

All learning done on the data sets using the two initialization approaches have been done with the same neighbourhood function and learning rate, this is to ensure that the two methods are subject to the same conditions which could influence the learning outcome of our study. To marginalize the effect of the sequence of training vectors, the applet adopts the batch learning SOM algorithm [10][12][4][5] described in detail in the next section. For the random initialization approach, the space of initial starting weights was sampled; this is because as the size of the data (n) increases, the possible initial starting configuration for a given K (the number of nodes) becomes quite enormous ($n^k$). The PCI was done using regular grid on the first principal component with equal variance [8]. For each data set and initialization approach, the data set was trained using three or four different values of K. Also the idea of quasi-linear model was presented.

## 2 Background

2.1 SOM Algorithm
The SOM is an artificial neural network which has a feed-forward structure with a single computational layer. Each neuron in the map is connected to all the input nodes.
   The on-line SOM algorithm as proposed by Kohonen can be summarised as follows:
   1) Initialization: An initial weight is assigned to all the connection. $W_j(0)$
   2) Competition: all nodes compete for the ownership of the input pattern. Using the Euclidean distant as criterion, the neuron with the minimum-distance wins.

$$J^* = \arg\min_k \| x(k) - w_j \| \quad ,j = 1,\ldots,MN,$$

   where $x(k)$ is the input pattern at time $k$
   3) Cooperation: the winning neuron also excites its neighbouring neurons (topologically close neurons). An example of neighbourhood function often used is the Gaussian neighbourhood function,

$$\eta_{*i}(k) = \alpha(k)\exp\left(-\frac{\| r_* - r_i \|^2}{2\sigma^2(k)}\right),$$

   where $\alpha(k)$ is a monotonically decreasing learning factor at time $(k)$, $r_i$ is the position of node $i$.
   4) Learning Process (Adaptation): The winning neuron and the neighbours are adjusted with the rule given below:

$$w_j(k+1) = w_j(k) + \alpha(k)\eta_{*j}(k)[x(k) - w_j(k)].$$

   Hence, the weight of the winning neuron and its neighbours are adjusted towards the input patterns however the neighbours have their weights adjusted with a value less than the winning neuron. This action helps to preserve the topology of the map.
   As $k \to \infty$, $\eta_{*i}(k) \to 0$.

2.3 The Batch Algorithm
   We use the batch algorithm of the SOM learning. This is a variant of the SOM algorithm in which the whole training set is presented to the map before the weights are adjusted with the net effect over the samples. The algorithm is given below [13].

   1) Present an input vector and find the winner neuron, which is the weight vector closest to the input data.

$$C_j = \arg\min_{i}\{\| w_i - x_j \|_2\}$$

2) Repeat step 1 for all the data points in the training set.
3) Update all the weights as follows

$$w_i(k+1) = \frac{\sum_{j=1}^{n} \eta_{ic(j)}(k) x_j}{\sum_{j=1}^{n} \eta_{ic(j)}(k)},$$

where $\eta_{ic(j)}$ is the neighbourhood function around the winner $c_j$ and $k$ is the number of iteration.

2.3 SOM learning algorithm used by the applet

Before learning, all $C_i$ are set to the empty set ($C_i=\emptyset$), and the steps counter is set to zero.

1. Associate data points with nodes (form the list of indices $C_i=\{l:\|x_l-y_i\|^2\leq\|x_l-y_j\|^2 \; \forall i\neq j\}$).
2. If all sets $C_i$, evaluated at step 1 coincide with sets from the previous step of learning, then STOP.
3. Add a new record to the learning history to place new coding vectors locations.
4. For every node, calculate the sum of the associated data points: $z_i=\sum_{j\in C_i} x_j$. If $C_i=\emptyset$ then $z_i=0$.
5. Evaluate new locations of coding vectors.
    1. For each node, calculate the weight $W_i=\sum h_{ij}|C_j|$ (here, $h_{ij}$ is the value of the neighbourhood function.)
    2. Calculate the sums $Z_i=\sum h_{ij} z_j$.
    3. Calculate new positions of coding vectors: $y_i v \leftarrow Z_i/W_i$ if $W_i\neq 0$. If $W_i=0$ then $y_i$ does not change.
6. Increment the step counter by 1.
7. If the step counter is equal to 100, then STOP.
8. Return to step 1.

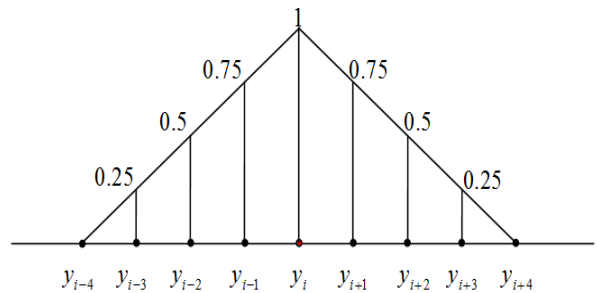

Fig. 1. The B-spline neighbourhood function with $h_{max}=3$.

The neighbourhood function used for this applet has the simple B-spline form given as a B-spline with $h_{max}=3$ (Fig. 1): $1-h_{ij}=|i-j|/(h_{max}+1)$ if $|i-j|\leq h_{max}$ and $h_{ij}=0$ if $|i-j|>h_{max}$

## 2.4 Fraction of Variance Unexplained.

For this applet, data are approximated by broken lines (SOM) [8]. The dimensionless least square evaluation of the error is the *Fraction of Variance Unexplained* (FVU). It is defined as the fraction: [The sum of squared distances from data to the approximating line /the sum of squared distances from data to the mean point].

The distance from a point to a straight line is the length of a perpendicular dropped from the point to the line. This definition allows us to evaluate FVU for PCA. For SOM we need to solve the following problem. For the given array of coding vectors $\{y_i\}$ ($i=1,2, ... m$) we have to calculate the distance from each data point $x$ to the broken line specified by a sequence of points $\{y_1, y_2, ... y_m\}$. For the data point $x$, its projection onto the broken line is defined, that is, the closest point. The square of distance between the coding vector $y_i$ and the point $x$ is $d_i(x)=\|x-y_i\|^2$ ($i=1,2, ... m$).

Let us calculate the squared distance from the data point $x$ to the segment $[y_i, y_{i+1}]$ ($i=1,2, ... m$-1). For each $i$, we calculate projection of a data point onto a segment. $l_i(x)=(x-y_i, y_{i+1}-y_i)/\|y_{i+1}-y_i\|^2$ (see figure 2).

If $0<l_i(x)<1$ then the point, nearest to $x$ on the segment $[y_i, y_{i+1}]$, is the internal point of the segment. Otherwise, this nearest point is one of the segment's ends.

Let $0<l_i(x)<1$ and $c$ be a projection of $x$ onto segment $[y_i, y_{i+1}]$. Then $\|c - y_i\|^2=(l_i(x)\|y_i-y_{i+1}\|)^2$ and, due to Pythagorean theorem, the squared distance from $x$ to the segment $[y_i, y_{i+1}]$ is $r_i(x)=\|x-y_i\|^2-(l_i(x) \|y_i-y_{i+1}\|)^2$.

Let $d(x)=\min\{d_i(x) \mid i=1,2, ... m\}$ and $r(x)=\min\{r_i(x) \mid 0< l_i(x) <1, 0< i< m \}$. Then the squared distance from $x$ to the broken line specified by the sequence of points $\{y_1, y_2, ... y_m\}$ is $D(x)=\min\{d(x), r(x)\}$.

For the given set of the data points $x_j$ and the given approximation by a broken line, the sum of the squared distances from the data points to the broken line is $S=\sum_j D(x_j)$, and the fraction of variance unexplained is $S/V$, where $V=\sum_j(x_j - X)^2$ and $X$ is the empirical mean: $X=(1/n)\sum x_i$.

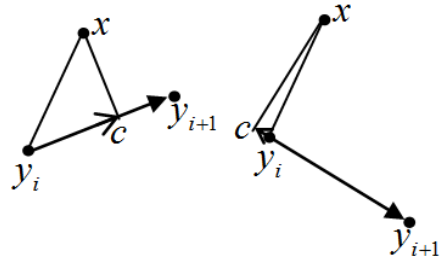

Fig. 2. A distance from a point to a segment: two versions of the projection.

## 2.5 Initialization Methods

As earlier stated, the objective of this paper is to consider the performance of two different initialization methods for SOM using the fraction of variance unexplained as the criterion for measuring the performance or quality of learning. The two initialization methods compared are:

- <u>PCA initialization (PCI)</u>: The weight vectors are selected from the subspace spanned by the first n principal component. And for this study the weight vectors are chosen as a regular grid on the first principal component with equal variance. Therefore given the number of weight vectors K, the behaviour of SOM using PCA initialization as described above is completely deterministic and also results in only one configuration. PCA initialization does not take into account the distribution of linear projection result and could produce several empty cells leading to the need for a reconstitution algorithm [1]. However according to [13], since the initialization is approximately organized, SOM computation can be made order of magnitude faster.
- <u>Random Initialization (RI)</u>: This method is very easy to implement, *K* weight vectors are selected randomly from the data points. Hence the possible size of initial starting configuration given a dataset of size n is given by $n^k$. However, given an initial configuration, the behaviour of the SOM becomes completely deterministic.

2.6 Linear, Quasi-Linear and Non-Linear models.

Data sets can be modelled using linear or non-linear manifold of lower dimension. According to [5] a class of non-linear data set was identified which could be modelled by quasi-linear model. In this study, data sets will be classified as quasi-linear or non-linear. To determine if a linear or non-linear model is appropriate for modelling a data set, see the non-linearity test for PCA [6]

**Linear Model** - data set is said to be linear if it could be modelled using a sequence of linear manifolds of small dimension (in 1D case, they can be approximated by a straight line with sufficient accuracy). These data can be easily approximated by the principal components. This class of data was not studied in this paper.

**Quasi-linear Model** – A dataset is called quasi-linear [5] if the principal curve approximating the dataset can be univalently projected on the linear principal component. It should be noted that the principal manifold is projected to the lines and not the nodes. For this study, dataset which falls in the border between non-linear and quasi-linear with in which over 50% of the data can be classified as quasi-linear is also classified as quasi-linear. See examples in figure 3a and 3c.

**Non linear Model** – For the purpose of this paper, essentially nonlinear datasets which do not fall into the class of quasi-linear datasets will be classified simply as nonlinear. See example in figure 3b.

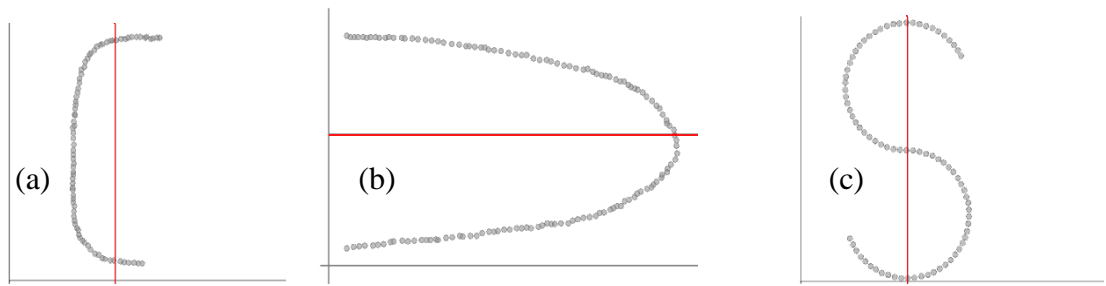

Fig. 3. (a) Quasi-linear data set; (b) nonlinear data set; (c) a border case between non-linear and quasi-linear dataset. The first principal component approximations are shown (red online).

# 3 Experimental Results

Using the SOM applet, the performance of both initialization methods on datasets with data distributed along different shapes (see figure 4) was studied at values of K = 10, 20, 50 (unless otherwise stated).

### 3.1 *Drawing up the Probability Distribution of FVU*
For the PCI as mentioned earlier, its yield just one initial configuration given K (this is because equidistant nodes are selected from the subspace of principal component such that the variances are equal).

In drawing up the probability distributions for the RI method, a sample of 100 initial configurations from the space of possible initial configurations for each dataset and each value of K was taken and the resulting FVU computed. The probability distribution of the FVU was described in terms of mean, median, standard deviation, minimum and maximum (see Table 3 in Appendix).

### 3.2 *Results*
As observed from the histogram (figure 5), and also the result of normality check carried out on the distributions, most of the distribution are not normal except for a few.

In order to compare the performance of the two initialization approaches in terms of the FVU, we consider the proportion estimate of RI which performs better than PCI and draw inferences. This is because for a given value of K, the PCI has only one configuration while the random method has a distribution. Therefore, from the distribution of FVU for RI, the proportion of FVU with values that are smaller than that of PCI (i.e. the proportion of RI in which the SOM map outperformed PCI in terms of FVU) was recorded and an interval estimate of the population proportion (using the adjusted Wald method) at confidence level of 95% was also computed. However, for cases where the sample proportion are extreme (i.e. proportion close to 0 or 100), the point estimate using the Laplace method was calculated. See Table 1

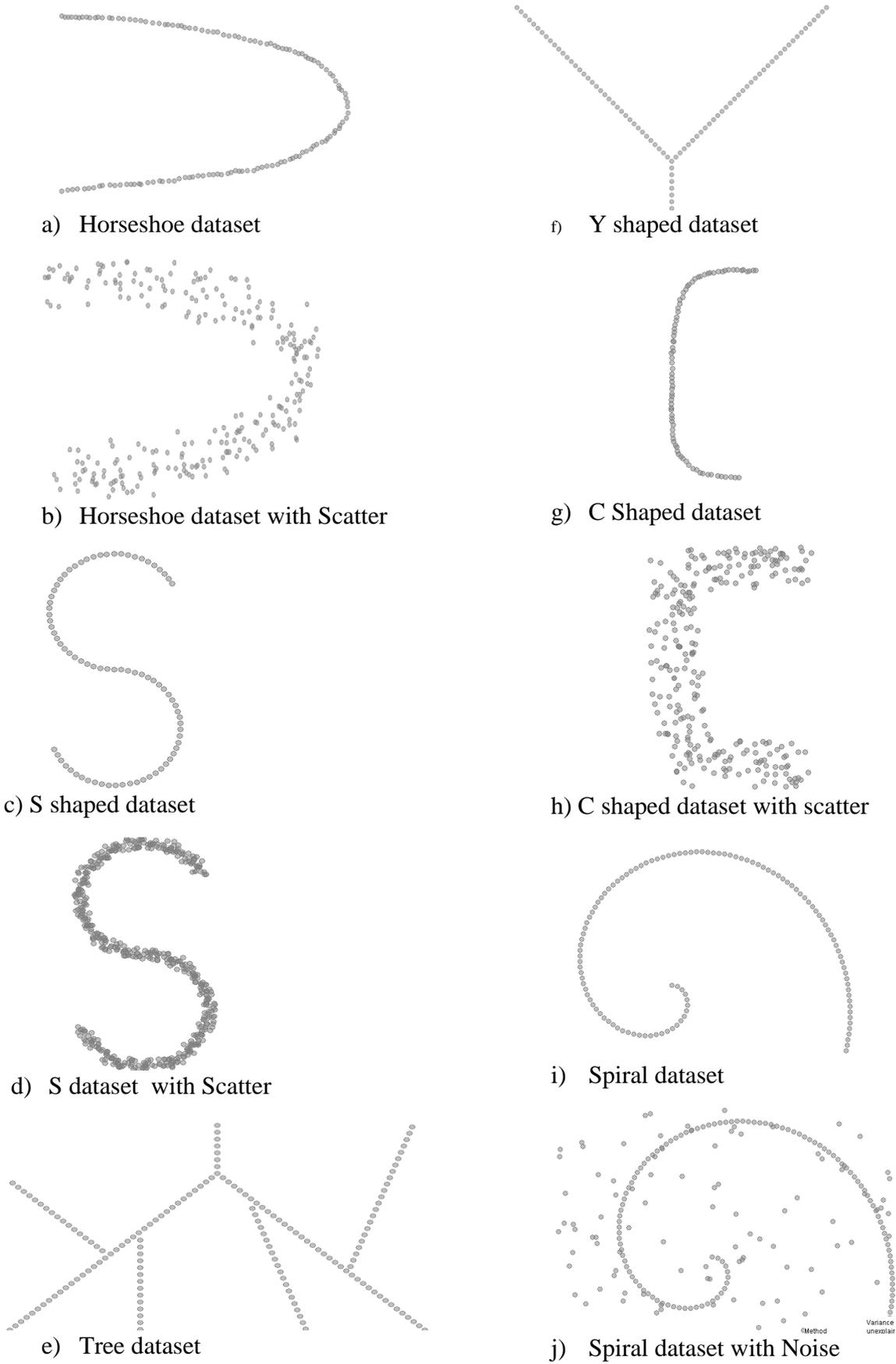

**Fig. 4. (a-j) Data sets for the case study.**

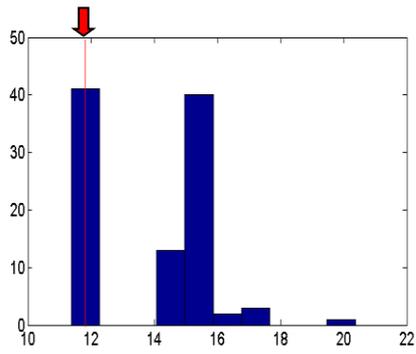
(a1) SOM approximation using 10 Nodes

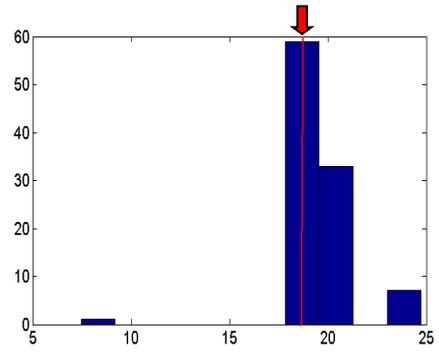
(b1) SOM approximation using 10 Nodes

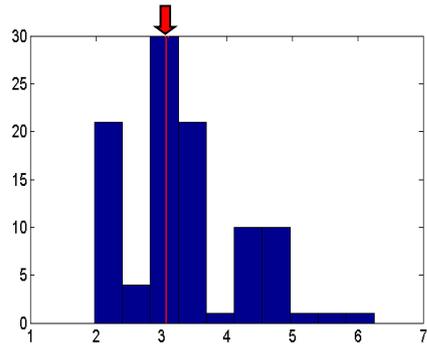
(a2) SOM approximation using 20 Nodes

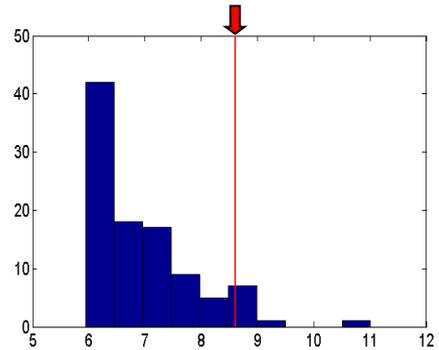
(b2) SOM approximation using 20 Nodes

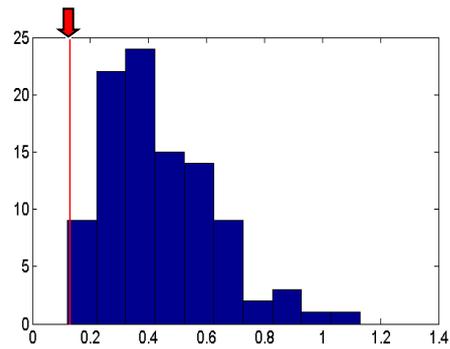
(a3) SOM approximation using 50 Nodes

**Fig 5(a1-a3). Spiral Data Set**

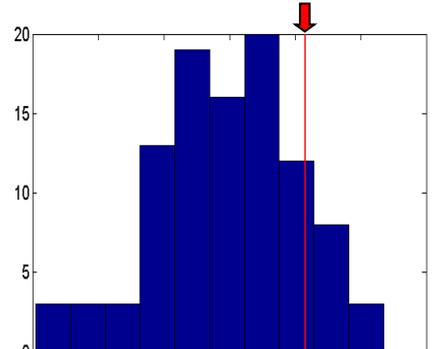
(b3) SOM approximation using 50 Nodes

**Fig 5(b1-b3). Spiral Data Set with Noise**

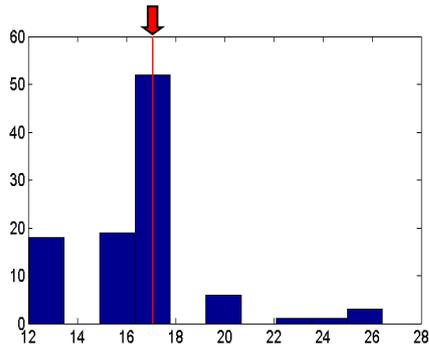

(c1) SOM approximation using 10 Nodes

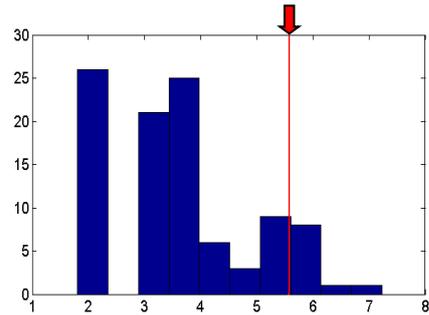

(c2) SOM approximation using 20 Nodes

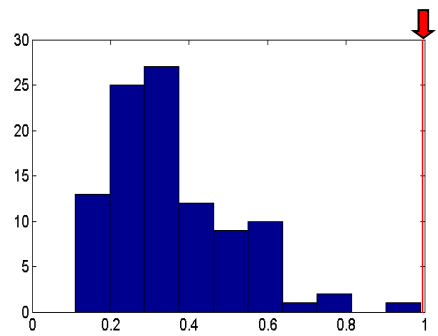

(c3) SOM approximation using 50 Nodes

**Fig 5(c1-c3). Horse Shoe Data Set**

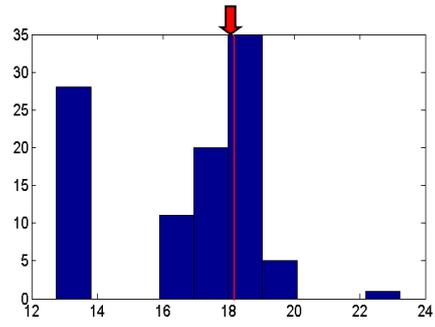

(d1) SOM approximation using 10 Nodes

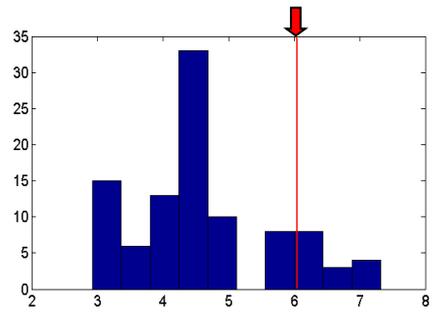

(d2) SOM approximation using 20 Nodes

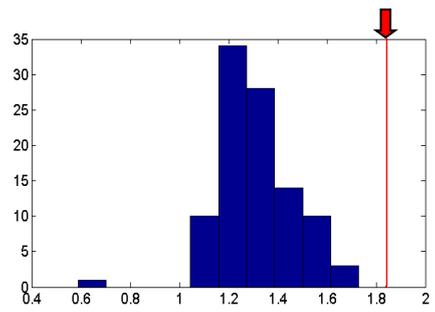

(d3) SOM approximation using 50 Nodes

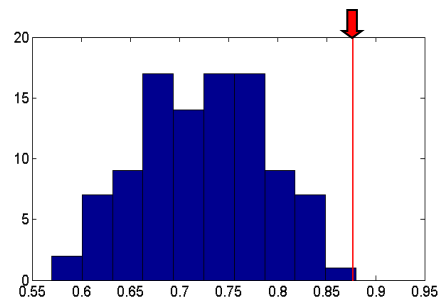

SOM approximation using 100 Nodes

**Fig 5(d1-d4). Horse Shoe Data Set with Scatter**

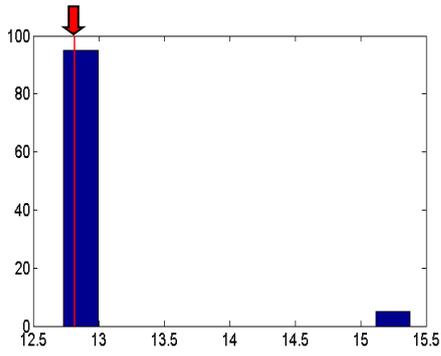

(e1) SOM approximation using 10 Nodes

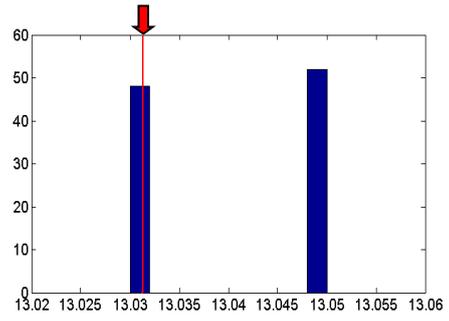

(f1) SOM approximation using 10 Nodes

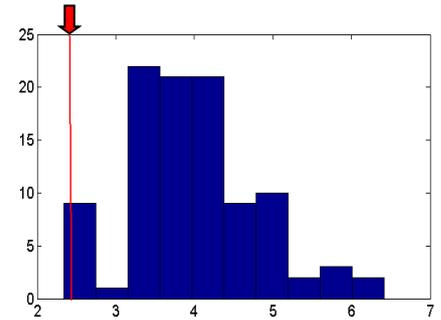

(e2) SOM approximation using 20 Nodes

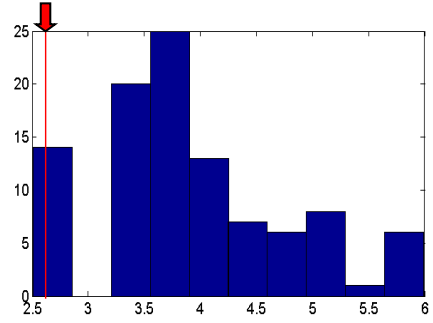

(f2) SOM approximation using 20 Nodes

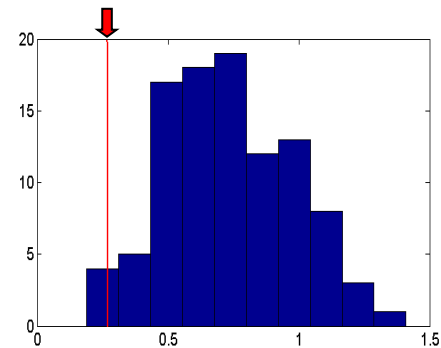

(e3) SOM approximation using 40 Nodes

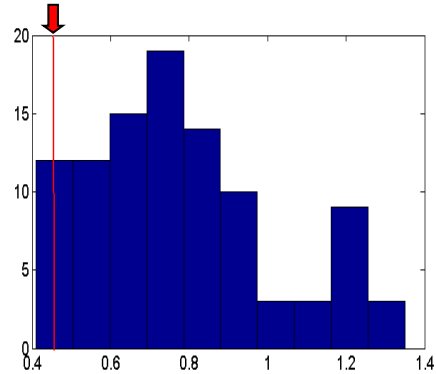

(f3) SOM approximation using 40 Nodes

**Fig 5(e1-e3). S shape Data Set**

**Fig 5(f1-f3). S shape Data Set with Scatter**

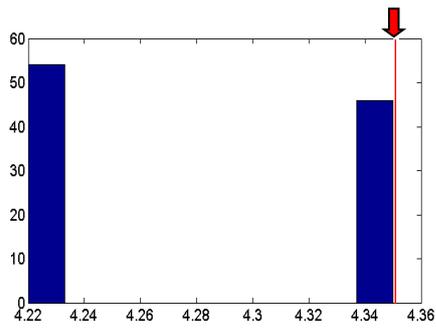

(g1) SOM approximation using 10 Nodes

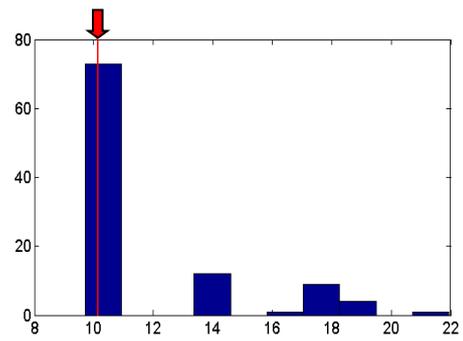

(h1) SOM approximation using 10 Nodes

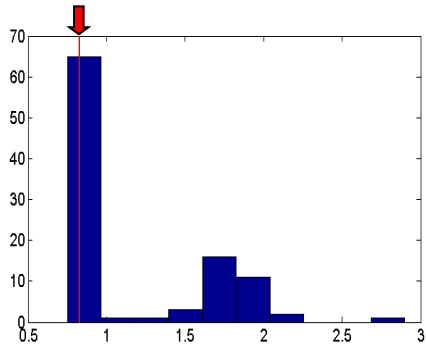

(g2) SOM approximation using 20 Nodes

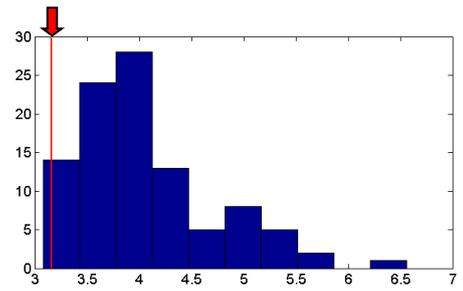

(h2) SOM approximation using 20 Nodes

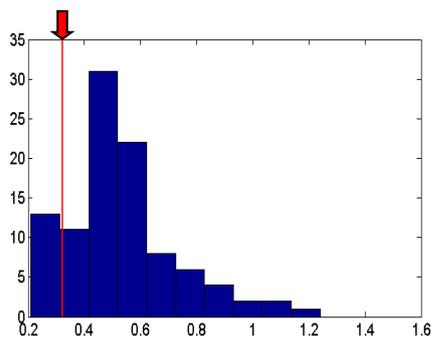

(g3) SOM approximation using 30 Nodes

**Fig 5g(1-3) C shape Data Set**

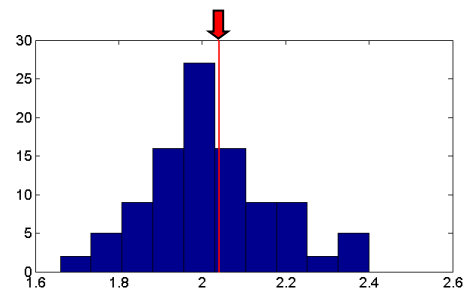

(h3) SOM approximation using 30 Nodes

**Fig 5h(1-3). C shape Data Set with scatter**

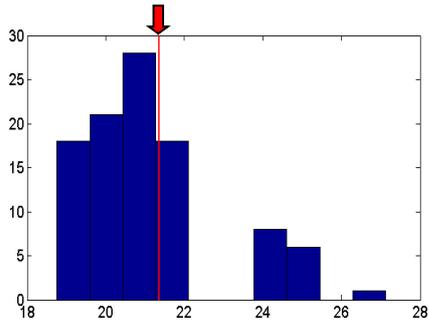
(i1) SOM approximation using 10 Nodes

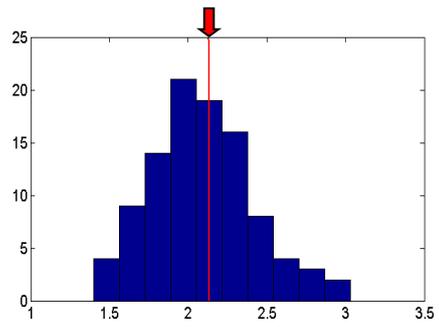
(i3) SOM approximation using 50 Nodes

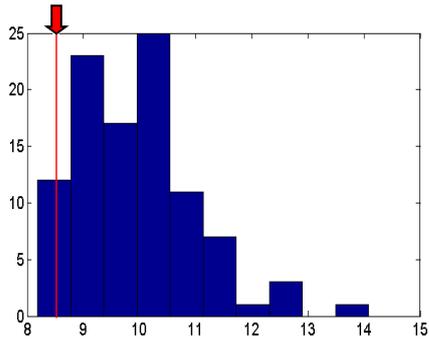
(i2) SOM approximation using 20 Nodes

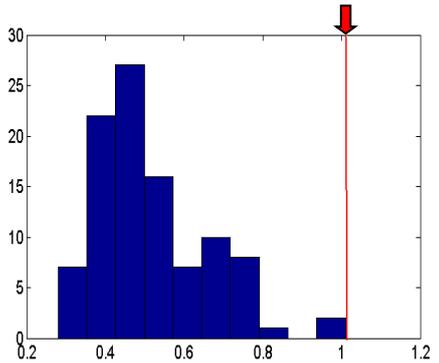
(i4) SOM approximation using 100 Nodes

**Fig 5i(1-4). Tree shape Data Set**

**Fig 5. Histograms of FVU for approximation of various datasets by 1D SOM with different numbers of nodes and Random Initiation. The values of FVU for the PCA initiations for the same dataset and number of nodes are highlighted by vertical lines and arrows ⬇(red online).**

**Table 1.** The proportion estimate of RI which performs better than PCI for various datasets and number of nodes

| Dataset | K | % better than PCI | Confidence Interval (%) at confidence level of 95% | Classification |
|---|---|---|---|---|
| Spiral 3 | 10 | 41% | 31.86 - 50.80 | Nonlinear |
|  | 20 | 55% | 45.24 - 64.39 | Nonlinear |
|  | 50 | 1% | 0 - 1.96 | Nonlinear |
|  |  |  |  |  |
| Spiral 3 With Noise | 10 | 49% | 39.42 - 58.65 | Nonlinear |
|  | 20 | 95% | 88.54 - 98.13 | Nonlinear |
|  | 50 | 84% | 75.47 - 90.01 | Nonlinear |
|  |  |  |  |  |
| Horse shoe | 10 | 73% | 65.53 - 80.77 | Nonlinear |
|  | 20 | 95% | 88.54 - 98.13 | Nonlinear |
|  | 50 | 100% | 99.02 - 100 | Nonlinear |
|  |  |  |  |  |
| Horse shoe with scatter | 10 | 74% | 64.58 - 81.64 | Nonlinear |
|  | 20 | 89% | 81.21 - 93.91 | Nonlinear |
|  | 50 | 100% | 99.02 - 100 | Nonlinear |
|  | 100 | 99% |  | Nonlinear |
|  |  |  |  |  |
| Tree | 10 | 76% | 66.71 - 83.36 | Nonlinear |
|  | 20 | 10% | 5.35 - 17.61 | Nonlinear |
|  | 50 | 67% | 57.28 - 75.46 | Nonlinear |
|  | 100 | 100% | 99.02-100 | Nonlinear |
|  |  |  |  |  |
| S Shape | 10 | 36% | 27.26 - 45.78 | Quasi-linear |
|  | 20 | 7% | 3.20 - 13.98 | Quasi-linear |
|  | 50 | 7% | 3.20 - 13.98 | Quasi-linear |
|  |  |  |  |  |
| S Shape with scatter | 10 | 48% | 38.46 – 57.68 | Quasi-linear |
|  | 20 | 9% | 4.62 – 16.42 | Quasi-linear |
|  | 50 | 11% | 6.09 - 18.79 | Quasi-linear |
|  |  |  |  |  |
| C Shape | 10 | 100% | 99.02-100 | Quasi-linear |
|  | 20 | 33% | 24.54 - 42.72 | Quasi-linear |
|  | 30 | 13% | 7.62 - 21.12 | Quasi-linear |
|  |  |  |  |  |
| C shape with scatter | 10 | 73% | 65.53 - 80.77 | Quasi-linear |
|  | 20 | 8% | 3.90 – 15.21 | Quasi-linear |
|  | 30 | 72% | 62.48 – 79.90 | Quasi-linear |

3.3 Analysis of Performance.

From the result above, it could be concluded that RI tend to perform quite well for nonlinear datasets. Very interesting result was obtained for the spiral dataset (Fig. 5a,b). For 10 nodes, 41% of RI realisations give better value of FVU than PCI, for 20 nodes this percent increases to 55% but for 50 nodes PCI gives better result than 99% of RI (Fig 5a3). We can conjecture that the SOM learning dynamics has many Stable Steady States SOMs (SSSSOMs). Sometimes the PCI can hit into a basin of attraction of a SSSSOM with low value of FVU. We have no different possible explanation of the surprising result presented in Fig 5a3.

It can be observed in Fig. 5b that the presence of noise affects the performance of the initialization methods. In particular, the surprisingly good performance of PCI of Fig 5a3 is destroyed by noise (Fig 5b3), and with noise the relative performance of PCI monotonically decreases with the number of nodes. In general, we can conclude that for essentially non-linear datasets PCI performs not better or even worse than the median of RI. The role of noise will be discussed later. However, the performance of RI is inconclusive regarding quasi-linear datasets. While the performance was good for the S shaped dataset the performance for the C shape was not as expected. For the approximation of the C shaped dataset by 1D SOM with 10 nodes all the results of RI were better than PCI gives (see Table 1). Nevertheless, it should be mentioned that the difference between the values of FVU for this case is rather small (Fig. 5g1). It does not exceed 4% of the minimal value of FVU (see Fig. 5g1). In that sense, the performance of PCI almost coincides with the quality of RI for example from Fig. 5g1.

Further analysis was performed to determine factors that could influence the performance of the initialization methods. By considering the effect of the under listed factors on the proportion of RI that outperforms PCI and using regression analysis the following were observed:

a) <u>Increase in Nodes (K)</u>: there was no relationship that could be established which indicates that increase in K significantly influence the performance of RI compared to PCI

b) <u>Number of unique final configuration in sample</u>: Even though the number of unique final configuration in the population is not well defined, there is a significant correlation between the number of unique final configurations in the sample and the performance of RI for quasi-linear datasets. This correlation however does not exist for non-linear data. See the result in table 2 below for quasi-linear datasets (and the raw data in Table 4 in Appendix).

Table 2. The correlation between the number of unique final configuration in the sample and the performance of RI for quasi-linear datasets in the case study.

**Coefficients[a]**

| Model | | Unstandardized Coefficients | | Standardized Coefficients | t | Sig. |
|---|---|---|---|---|---|---|
| | | B | Std. Error | Beta | | |
| 1 | (Constant) | .713 | .121 | | 5.911 | .000 |
| | Unique | -.010 | .003 | -.749 | -3.576 | .005 |

a. Dependent Variable: Proportion

c) <u>Increase in the data points (N)</u>: increase in N does not significantly influence the performance of RI compared with PCI.

d) <u>Presence of noise:</u> It was observed that the presences of noise in the spiral dataset tend to influence the performance of PCI. Further studies show that the presence of noise in quasi-linear data sets affects the performance of PCI. This is because noise can affect the principal component and also the principal curve of the dataset, which can affect the classification of dataset especially for quasi-linear datasets. An illustration is given in figure 6.

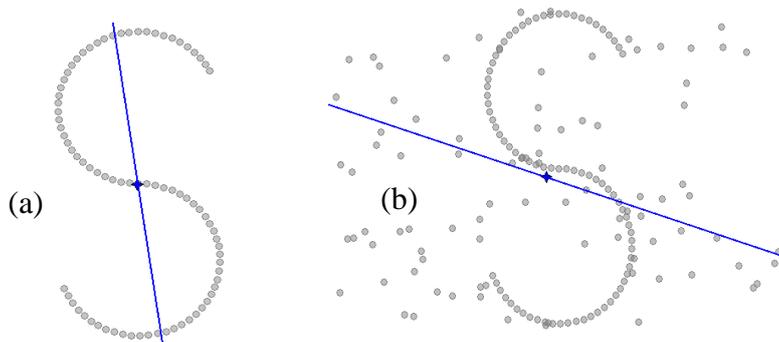

Fig. 6. (a) S shaped dataset (it is almost quasi-linear). S shape dataset with noise (it becomes essentially non-linear). The straight lines are the first principal component approximations (blue line).

## 4. Conclusion

It is widely known that the quality of SOM map is influenced by initial weight of the map. This study seeks to compare the performance of random initialization method and principal component initialization method using the SOM batch algorithm on datasets which has been classified as quasi-linear and non-linear. Datasets with data distributed along different shapes was studied and the performance was evaluated by fraction of variance unexplained. By comparing the proportion of final SOM map of RI which outperformed PCI under the same conditions, it was observed that RI performed quite well for non-linear data sets. However for quasi-linear datasets, the result remains inconclusive. In general, we can conclude that the hypothesis about advantages of the PCI is definitely wrong for essentially nonlinear datasets.

## References

[1] Attik, M.; Bougrain, L.; Alexandre, F: 'Self-organizing Map Initialization', Duch, W. ; Kacprzyk, J. ; Oja, E. ; Zadrozny, S., *Artificial Neural Networks: Biological Inspirations:* Lecture Notes in Computer Science. Volume 3696: Springer Berlin Heidelberg. Isbn: 978-3-540-28752-0, pp 357-362, 2005.
[2] Bullinaria, J.A.: 'Self Organizing Maps': Fundamentals, Introduction to Neural networks: Lecture 16. University of Birmingham, 2004
[3] Fort, J.-C.; Cottrell, M.; Letremy, P. 2001: 'Stochastic on-line algorithm versus batch algorithm for quantization and self organizing maps, Neural Networks for Signal Processing XI'. *Proceedings of the 2001 IEEE Signal Processing Society Workshop*, vol., no., pp.43-52, 2001.
[4] Fort, J-C. ; Letrémy, P. and Cottrell, M. 'Advantages and drawbacks of the batch Kohonen algorithm'. In M.Verleysen, editor, ESANN'2002 *Proceedings, European Symposium on Artificial Neural Networks,* Bruges (Belgium), pages 223–230, Bruxelles, Belgium, 2002. Editions D Facto.

[5] Gorban, A.N.; Rossiev, A.A.; Wunsch II, D.C: 'Neural network modeling of data with gaps: method of principal curves, Carleman's formula, and other', A talk given at the USA-NIS Neurocomputing opportunities workshop, Washington DC, July 1999 (Associated with IJCNN'99), arXiv:cond-mat/0305508 [cond-mat.dis-nn]

[6] Kruger, U.; Zhang, J.; Xie, L.: 'Development and Apllications of Nonlinear Principal Component Analysis – a Review'. In Gorban, A.N. ; Kegl, B. ; Wunsch, D.C. ; Zinovyev, A.Y., *Principal Manifolds for Data Visualization and Dimension Reduction*: Lecture Notes in Computational Science and Engineering. Volume 58: Springer Berlin Heidelberg. Isbn: 978-3-540-73750-6, pp 1-43, 2008.

[7] Matsushita, H.; Nishio, Y.: 'Batch-Learning Self-Organizing Map with false-neighbor degree between neurons," *Neural Networks, 2008. IJCNN 2008. (IEEE World Congress on Computational Intelligence). IEEE International Joint Conference on* , vol., no., pp. 2259-2266, 1-8 June 2008.

[8] Mirkes, E.M. Principal Component Analysis and Self-Organizing Maps: applet. University of Leicester, 2011. http://www.math.le.ac.uk/people/ag153/homepage/PCA_SOM/PCA_SOM.html

[9] Pena, J.M., Lozano, J.A. and Larranaga P.: 'An empirical comparison of four initialization methods for the K-Means algorithm'. *Pattern recognition Letters* 20, pp 1027-1040, 1999.

[10] Ghosh, A. P.; Maitra, R. and Peterson, A. D: 'A systematic evaluation of different methods for initializing the k-means clustering algorithm'. Technical Report 07, Iowa State University, Department of Statistics, Ames, IA, 2010.

[11] Su, M.-C.; Liu, T.-K.; and Chang, H.-T.:' Improving the Self-Organizing Feature Map Algorithm Using an Efficient Initialization Scheme', *Tamkang Journal of Science and Engineering*, Vol. 5, No. 1, pp. 35-48, 2002

[12] Kohonen, T., Honkela, T. Kohonen network. Scholarpedia, 2(1):1568., revision #122029, 2007.

[13] Kohonen T.: The Self-Organizing Map (SOM), 2005. accessed 20/07/12. http://www.cis.hut.fi/projects/somtoolbox/theory/somalgorithm.shtml

[14] Yin, H.: 'The Self-Organizing Maps: Background, Theories, Extensions and Applications'. In Fulcher, J. et al. (Eds), *Computational Intelligence: A Compendium: Studies in Computational Intelligence*, Springer Berlin / Heidelberg, pp 715-762, 2008.

[15] Yin H. Learning Nonlinear Principal Manifolds by Self-Organising Maps, In: Gorban A. N. et al. (Eds.), LNCSE 58, Springer, 2007.

# Appendix

Table 3. The distribution of FVU for RI. (The last column is FVU for PCI)

| Dataset | K | Mean | Std | Min | Max | PCI |
|---|---|---|---|---|---|---|
| Spiral 3 | 10 | 13.81 | 2.117 | 11.38 | 20.37 | 11.43 |
|  | 20 | 3.31 | 0.93 | 1.98 | 6.25 | 3.26 |
|  | 50 | 0.44 | 0.19 | 0.12 | 6.25 | 0.13 |
| Spiral 3 With Noise | 10 | 19.24 | 1.84 | 7.47 | 24.74 | 18.52 |
|  | 20 | 6.92 | 0.93 | 5.95 | 11.01 | 8.65 |
|  | 50 | 2.4 | 0.22 | 1.81 | 2.87 | 2.61 |
| Horse shoe | 10 | 16.51 | 2.86 | 12.03 | 26.43 | 17.12 |
|  | 20 | 3.55 | 1.33 | 1.81 | 7.23 | 5.99 |
|  | 50 | 0.35 | 0.16 | 0.11 | 0.993 | 1.32 |
| Horse shoe with scatter | 10 | 16.52 | 2.45 | 12.76 | 23.23 | 18.27 |
|  | 20 | 4.6 | 1.09 | 2.93 | 7.32 | 6.15 |
|  | 50 | 1.31 | 0.15 | 0.59 | 1.73 | 1.91 |
|  | 100 | 0.72 | 0.06 | 0.57 | 0.88 | 0.86 |
| Tree | 10 | 21.17 | 1.83 | 18.76 | 27.12 | 21.72 |
|  | 20 | 9.94 | 1.04 | 8.19 | 14.08 | 8.66 |
|  | 50 | 2.09 | 0.34 | 1.4 | 3.03 | 2.21 |
|  | 100 | 0.52 | 0.15 | 0.28 | 1.01 | 1.01 |
| S Shape | 10 | 12.89 | 0.54 | 12.73 | 15.38 | 12.76 |
|  | 20 | 3.96 | 0.84 | 2.34 | 6.42 | 2.37 |
|  | 50 | 0.73 | 0.25 | 0.19 | 1.41 | 0.35 |
| S Shape with scatter | 10 | 13.04 | 0.01 | 13.03 | 13.05 | 13.03 |
|  | 20 | 3.91 | 0.87 | 2.51 | 5.99 | 2.52 |
|  | 50 | 0.78 | 0.24 | 0.41 | 1.35 | 0.46 |
| C Shape | 10 | 4.28 | 0.07 | 4.22 | 4.35 | 4.35 |
|  | 20 | 1.19 | 0.48 | 0.75 | 2.9 | 0.88 |
|  | 30 | 0.53 | 0.19 | 0.21 | 1.24 | 0.31 |
| C shape with scatter | 10 | 11.41 | 3.05 | 9.7 | 21.94 | 9.78 |
|  | 20 | 4.04 | 0.67 | 3.08 | 6.56 | 3.13 |
|  | 30 | 2.02 | 0.15 | 1.66 | 2.4 | 2.07 |

Table 4. Number of unique final configuration in the sample
and the relative performance of RI versus PCI
for quasi-linear datasets in the case study.

| % of PCI greater than RI | Unique final configuration |
|---|---|
| 100.00% | 2 |
| 33.00% | 34 |
| 13.00% | 47 |
| 73.00% | 11 |
| 8.00% | 50 |
| 72.00% | 47 |
| 36.00% | 11 |
| 7.00% | 67 |
| 7.00% | 65 |
| 48.00% | 2 |
| 9.00% | 49 |
| 11.00% | 58 |